%% file: multimedia.tex
\documentclass[journal]{IEEEtran}

\ifCLASSINFOpdf
  \usepackage[pdftex]{graphicx}
  \graphicspath{{figures/}{pictures/}{images/}{./}}
  \DeclareGraphicsExtensions{.pdf,.jpeg,.png,.jpg}
\else
  \usepackage[dvips]{graphicx}
  \graphicspath{{../eps/}}
  \DeclareGraphicsExtensions{.eps}
\fi

\usepackage{amsmath}
\usepackage{amssymb} 
\usepackage{balance} 

\usepackage{url}
\usepackage{booktabs} 
\usepackage{enumitem}
\setlist{nosep}

\usepackage[hidelinks]{hyperref}
\usepackage{orcidlink}

\usepackage{pifont}
\newcommand{\cmark}{\ding{51}}%
\newcommand{\xmark}{\ding{55}}%

\usepackage[table]{xcolor}
\usepackage{colortbl}
\definecolor{lightgray}{gray}{0.9}

\begin{document}
%
\title{GenKOL: Modular Generative AI Framework For Scalable Virtual KOL Generation}

\author{
    Tan-Hiep To\orcidlink{0009-0008-1264-1354} \textsuperscript{1,2}
    , Duy-Khang Nguyen\orcidlink{0009-0002-2207-5065}\textsuperscript{1,2}
    , Tam V. Nguyen\orcidlink{0000-0003-0236-7992}\textsuperscript{3},
    Minh-Triet Tran\orcidlink{0000-0003-3046-3041}\textsuperscript{1,2}, 
    Trung-Nghia Le\orcidlink{0000-0002-7363-2610}\textsuperscript{1,2}*\thanks{*Corresponding author. Email: ltnghia@fit.hcmus.edu.vn} \\ 
    \textsuperscript{1}University of Science, VNU-HCM, Ho Chi Minh City, Vietnam \\
    \textsuperscript{2}Vietnam National University - Ho Chi Minh, Ho Chi Minh City, Vietnam \\
    \textsuperscript{3}University of Dayton, Ohio, US 
}


\markboth{Journal Submission}%
{T.-H. To \MakeLowercase{\textit{et al.}}: GenKOL: Modular Generative AI Framework For Scalable Virtual KOL Generation}

\maketitle

\begin{abstract}
Key Opinion Leader (KOL) play a crucial role in modern marketing by shaping consumer perceptions and enhancing brand credibility. However, collaborating with human KOLs often involves high costs and logistical challenges. To address this, we present GenKOL, an interactive system that empowers marketing professionals to efficiently generate high-quality virtual KOL images using generative AI. GenKOL enables users to dynamically compose promotional visuals through an intuitive interface that integrates multiple AI capabilities, including garment generation, makeup transfer, background synthesis, and hair editing. These capabilities are implemented as modular, interchangeable services that can be deployed flexibly on local machines or in the cloud. This modular architecture ensures adaptability across diverse use cases and computational environments. Our system can significantly streamline the production of branded content, lowering costs and accelerating marketing workflows through scalable virtual KOL creation.
\end{abstract}



%
\IEEEpeerreviewmaketitle


\input{content/introduction}
\input{content/relatedwork}
\input{content/method}
\input{content/exp}
\input{content/conclusion}

\section*{Acknowledgment}

This research is funded by Vietnam National Foundation for Science and Technology Development (NAFOSTED) under Grant Number 102.05-2023.31.



\bibliographystyle{IEEEtran}
\bibliography{References}

%










\end{document}

%% file: content/introduction.tex
\section{Introduction}
\label{introduction}

Key Opinion Leaders (KOLs) are influential figures in specific domains and communities, particularly in marketing. Partnering with KOLs can enhance a brand’s reputation and strongly shape consumer perceptions of its products and services~\cite{Kol}. However, such collaborations also present notable challenges. Working with high-profile KOLs often requires substantial financial investment, placing considerable strain on marketing budgets~\cite{Kol}. Moreover, producing and managing the necessary content and visuals demands significant time and effort. These challenges underscore the need for cost-effective alternatives that preserve engagement quality while reducing resource expenditure.

The rapid progress of generative AI offers a promising solution: virtual Key Opinion Leaders (KOLs). By leveraging deep learning models, organizations have the ability to fully or partially automate the generation of marketing materials such as images, videos, synthetic audio, and highly customizable digital personas that appeal to their target audience~\cite{gen_models}. This approach not only reduces costs and production time but also provides unprecedented creative flexibility and narrative adaptability. In contrast, traditional image editing software requires extensive expertise and lengthy training, limiting its accessibility for non-expert users. These limitations further motivate the development of systems that can democratize high-quality content creation.

Despite advances in technology, existing AI models still have difficulty generating high-quality images that align with user expectations. These models can be complicated and may fail to accurately reflect users' intentions, which limits their ability to combine images flexibly. The integration of multiple deep learning models, particularly generative AI models~\cite{gen_models}, often demands substantial hardware and software resources, which hinders user accessibility and reduces flexibility and scalability. Thus, it is both essential and strategically beneficial to create a flexible system that facilitates the straightforward incorporation of various AI models using a plugin-based resource distribution approach. Developing an integrated application that is user-friendly, cost-effective, and efficient in time while producing high-quality results is vital. This would offer effective solutions not just for product advertising but also for a wide range of applications in commerce, marketing, and content creation.

\begin{figure}[!t]
    \centering
    \includegraphics[width=\linewidth]{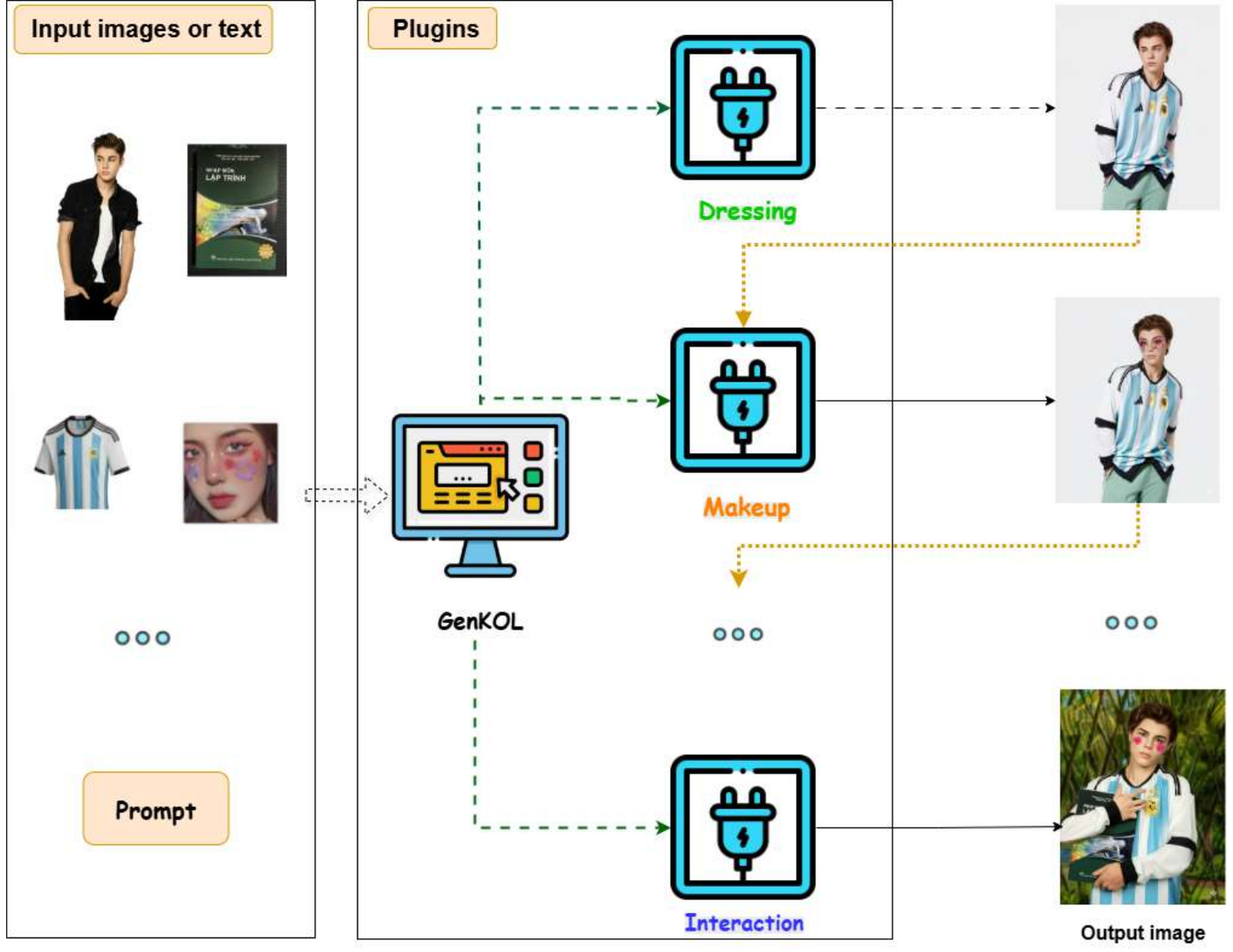}
    \caption{Workflow of the proposed GenKOL system.}
    \label{fig:kol_system}
\end{figure}

In this paper, we present GenKOL, a deep learning system for creating virtual KOLs that offers a scalable and effective solution for marketing and content production. GenKOL allows for efficient management of resources and scalable, modular workflows, facilitating the simple addition of new AI features without extensive modifications. Particularly, we introduce a plugin-driven framework that modularizes tasks such as garment generation~\cite{imagdressing, nguyen2023dm}, makeup transfer~\cite{stable-makeup}, and background synthesis~\cite{eshratifar2024salient, islam2024gemini}. Each component operates as an independent, deployable service that can run across heterogeneous environments, including local devices, cloud platforms. This design enables flexible updating, swapping, or scaling of AI services according to user requirements or computational resources. By abstracting model functionalities through standardized interfaces, the framework improves reusability, accelerates development, and ensures cross-platform compatibility. Figure~\ref{fig:kol_system} illustrates the system workflow, which is demonstrated at ACM Multimedia 2025~\cite{genkol}.


Our main contributions are as follows:
\begin{itemize}
    \item We introduce a generative AI framework that enables seamless integration of a base identity image with multiple stylistic references, while preserving semantic consistency and maintaining distinct visual characteristics.
    
    \item We design a modular and extensible system architecture that facilitates the integration of new algorithms and services. This design enhances adaptability to evolving technologies and empowers users to tailor the generative pipeline for diverse, application-specific requirements.
\end{itemize}


%% file: content/relatedwork.tex
\section{Related Work}
\label{related}

Image generation has become a central topic in artificial intelligence, fueled by advances in deep learning. Early approaches were dominated by Generative Adversarial Networks (GANs)~\cite{gans}, which introduced an adversarial training framework capable of producing realistic, high-quality images. GAN-based methods have been successfully applied to diverse tasks, including image-to-image translation, sketch-to-image synthesis, conditional and text-guided generation, video synthesis, panoramic rendering, and scene graph–based generation.

More recently, diffusion models~\cite{diffusion_models_1} have emerged as a powerful alternative, offering improved training stability, sample diversity, and controllability. These models generate images by progressively corrupting training data with Gaussian noise (forward process) and learning to invert this corruption (reverse process). By optimizing a variational lower bound on data likelihood, diffusion-based approaches achieve precise control and consistently higher quality outputs. Foundational contributions by Ho et al.~\cite{diffusion_models_1} and Rombach et al.~\cite{stable-diffusion-2} established the scalability and visual fidelity of diffusion models. Subsequent studies expanded their applicability: Tumanyan et al.~\cite{diffusionfeatures} explored feature-level control, Baranchuk et al.~\cite{author_baranchuk} investigated conditional sampling, and Xu et al.~\cite{author_xu} advanced semantic image synthesis. Building on these developments, our work leverages state-of-the-art generative models~\cite{gen_models} to create expressive, customizable, and high-quality virtual KOL images.

\begin{figure}[!t]
\centering
\includegraphics[width=\linewidth]{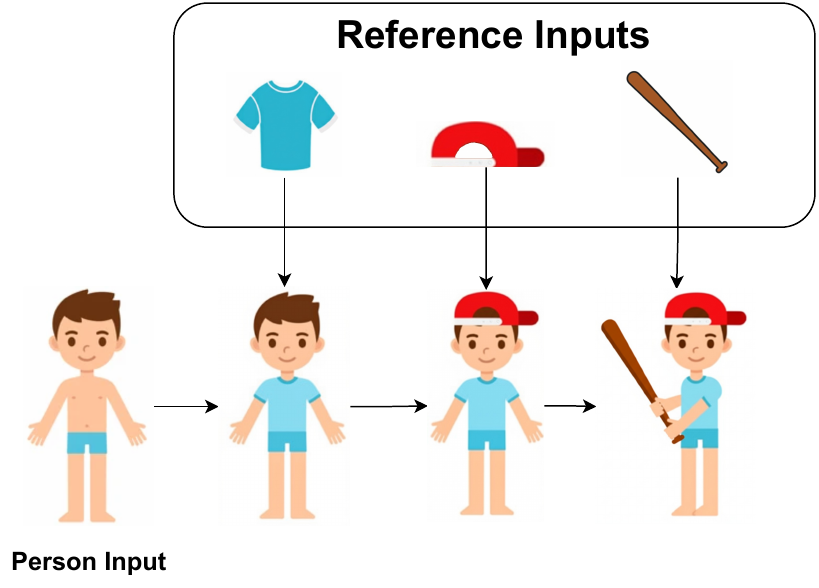}
\caption{Illustration of multistep image generation pipeline.}
\label{fig:multistep}
\end{figure}

Another emerging direction is multi-step image generation, which decomposes synthesis into a sequence of refinement operations. Instead of producing an image in a single pass, multi-step pipelines allow iterative editing, enabling fine-grained personalization. This paradigm is especially relevant for applications such as avatar creation and interactive visual editing, where user control is critical (Figure~\ref{fig:multistep}). Modern systems such as ControlNet~\cite{controltexttoimage} and PhotoMaker~\cite{photomakercustomizin} exemplify this approach by supporting staged transformations—adjusting human pose, editing backgrounds, or modifying facial attributes. While multi-step pipelines enhance interpretability and user control by exposing intermediate outputs, they also introduce challenges such as error accumulation, increased complexity, and the need for standardized interfaces between modules.

In GenKOL, \textbf{multi-step generation is a core design principle}. Each service functions as an independent generative unit that consumes both reference inputs and intermediate outputs to produce refined results for subsequent stages. This modular structure aligns with the broader trend toward flexible, user-centric generative systems, while addressing key limitations in scalability, adaptability, and usability.

%% file: content/method.tex
\section{Proposed GenKOL System}
\label{method}

\subsection{Overview}

\begin{figure*}[!t]
    \centering
    \includegraphics[width=\textwidth]{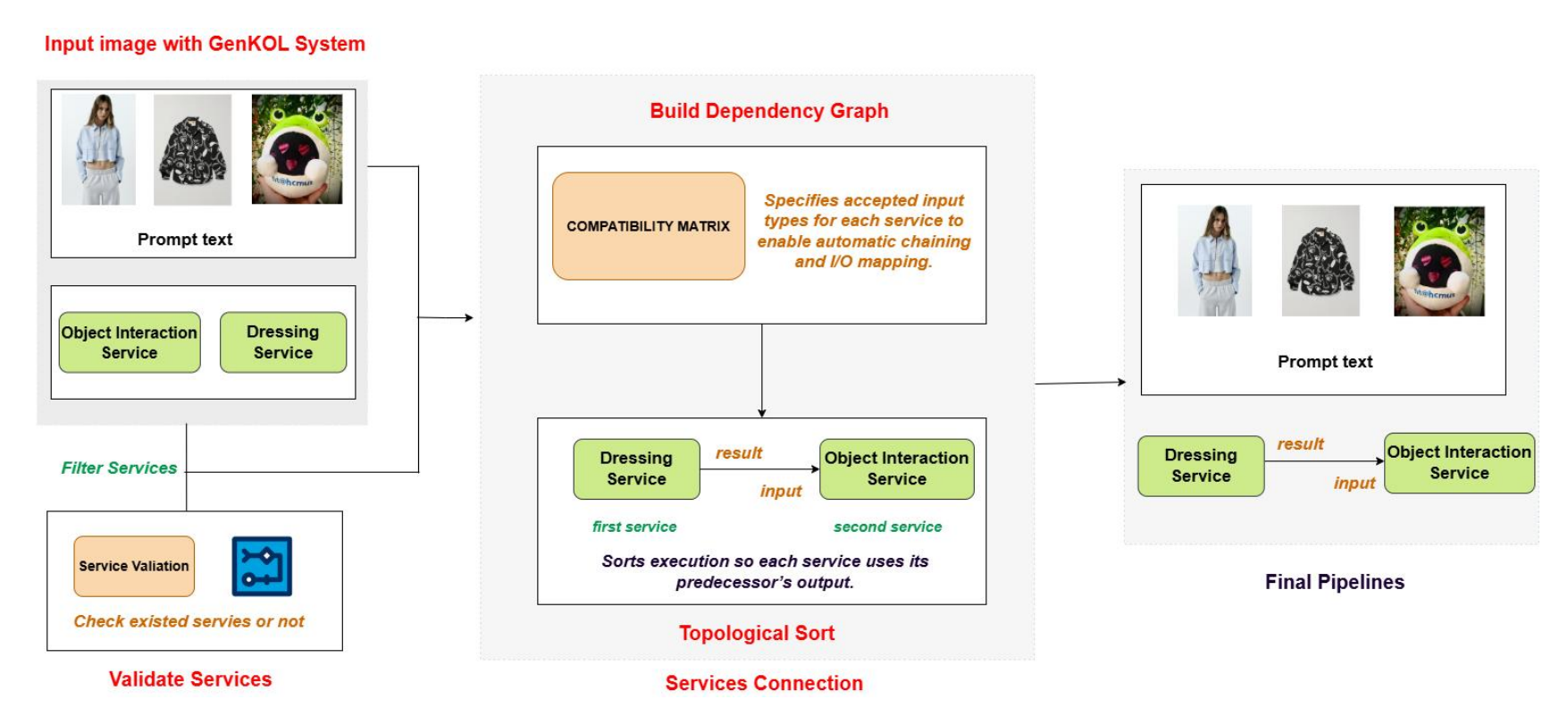}
    \caption {Compatibility and intelligent input–output mapping for pipeline generation.}
    \label{fig:service_mapping_workflow}
\end{figure*}

\begin{figure}[t!]
    \centering
\includegraphics[width=\linewidth]{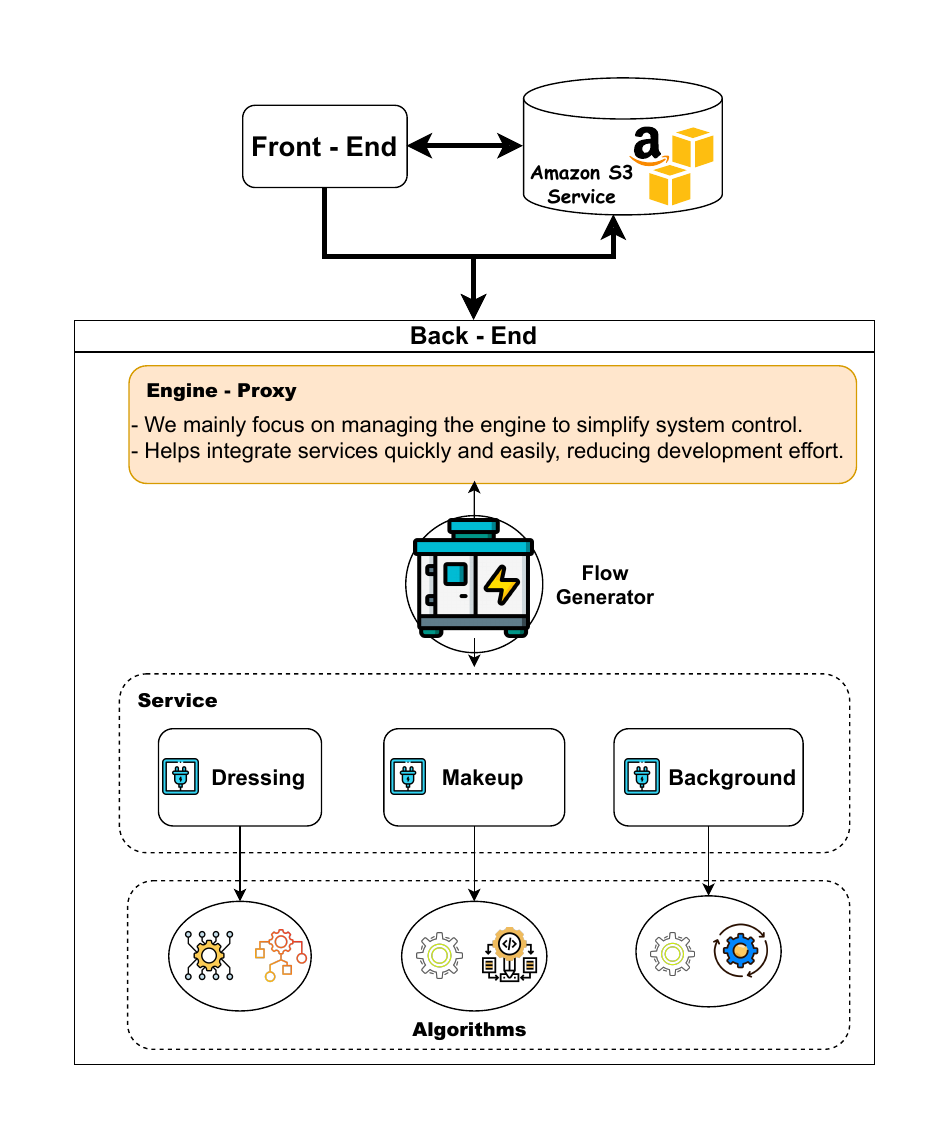}
    \caption{Proposed modular AI services architecture.}
    \label{fig:proposed_system}
\end{figure}

GenKOL utilizes a flexible and modular AI plugin architecture, as shown in Figure~\ref{fig:proposed_system}. This allows users to create context-rich virtual KOL visuals from input images and detailed text prompts while maintaining strong contextual coherence. As a result, it outperforms traditional, loosely integrated pipelines. The system's modular design supports scalable deployment, efficient use of resources, and seamless integration of various generative models. By chaining specialized services into customized workflows, users can ensure smooth data flow, accelerate experimentation, and consistently produce high-quality, contextually aligned outputs.

A central principle of the system is the establishment of clear execution pathways among modular AI services. Tasks such as outfit replacement, virtual makeup, background editing, and object interaction are connected in a logical sequence. To streamline execution, GenKOL employs an automated orchestration process that determines the appropriate ordering and connections among services. Specifically, topological sorting is applied to organize services as a directed acyclic graph (DAG), ensuring that no service executes before its prerequisites are satisfied. A compatibility matrix further validates potential connections, preventing incompatible service pairings and avoiding execution failures. Once validated, each service is automatically assigned the required inputs from preceding outputs, thereby supporting flexible and reliable workflow construction.

\begin{figure}[!t]
    \centering
    \includegraphics[width=\linewidth]{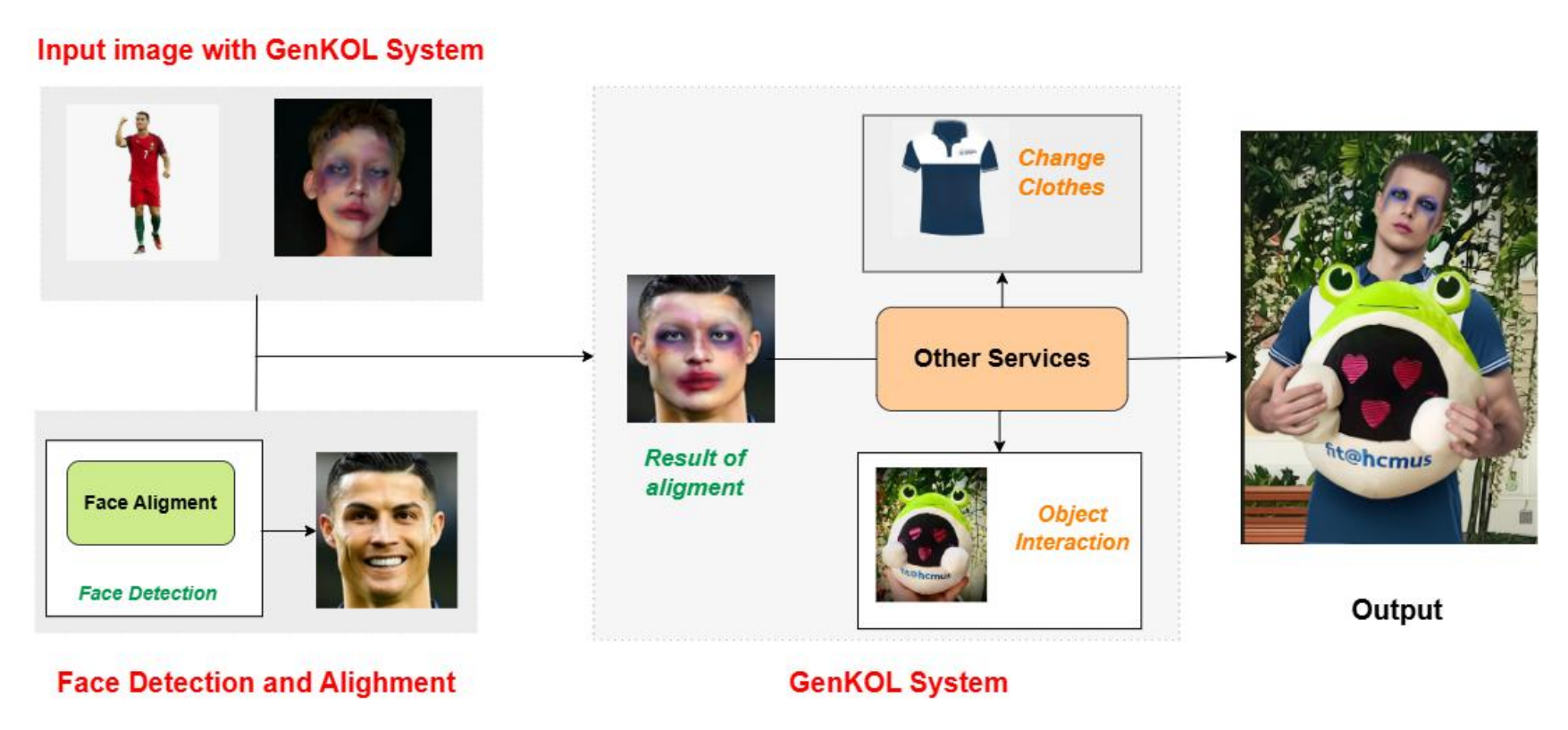}
    \caption{Overview of the face detection and alignment procedure using a pretrained landmark model, applied to ensure pose normalization before executing GenKOL's generative services.}
    \label{fig:kol_face_extractor}
\end{figure}

Figure~\ref{fig:service_mapping_workflow} demonstrates the comprehensive workflow for verifying compatibility and integrating services within the GenKOL pipeline. Its intelligent mapping system automates data transfer between services, minimizing manual intervention and reducing configuration errors. This automation allows users to easily create complex workflows directly from search queries while maintaining flexibility, technical correctness, and proactive error management, ultimately enhancing modularity, scalability, and dynamic deployment.

\begin{figure*}[!t]
    \centering
\includegraphics[width=\textwidth]{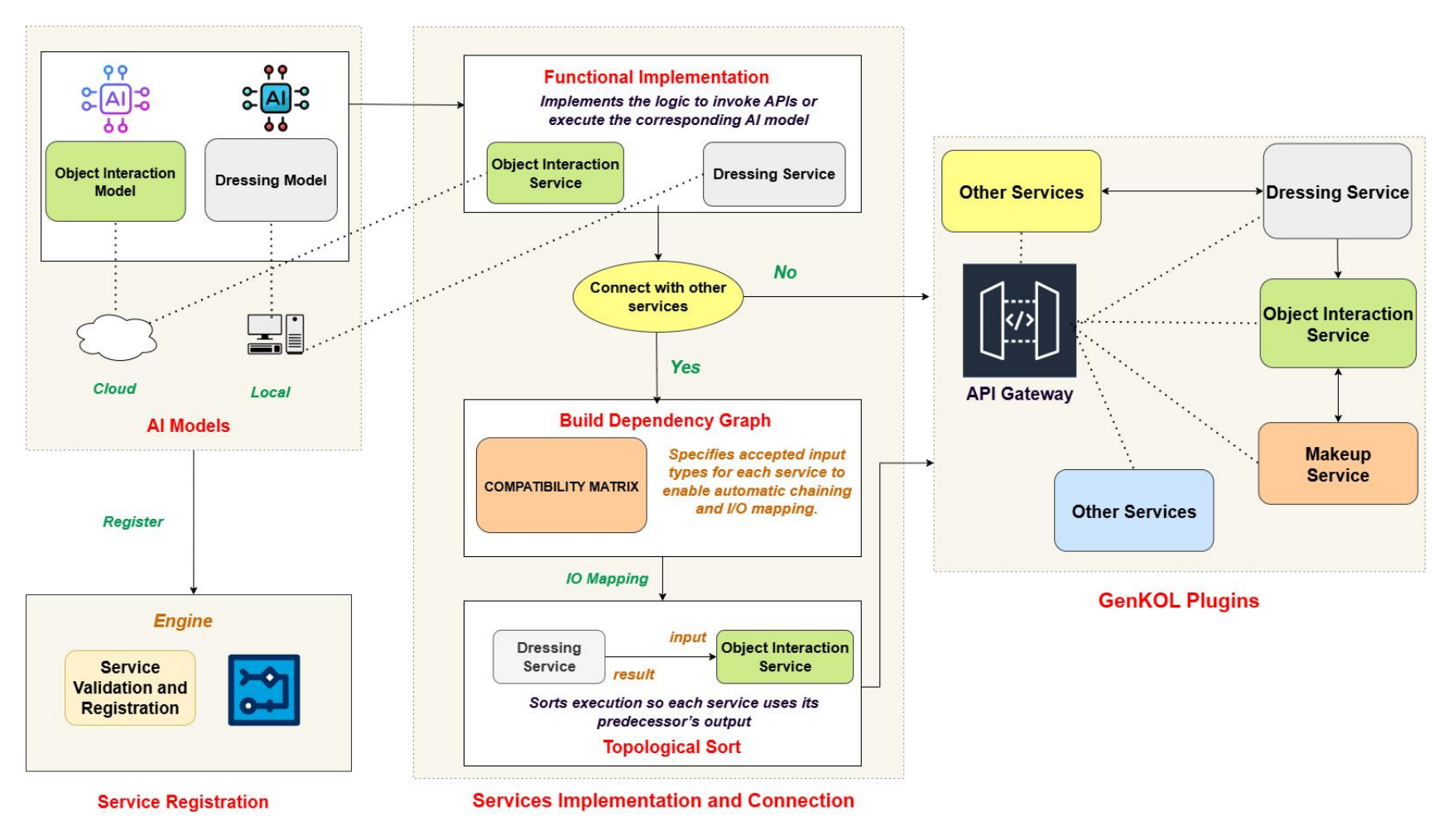}
    \caption{Overview of the plugin-based integration workflow in GenKOL.}
    \label{fig:modular-extend}
\end{figure*}

A key challenge arises from variations in facial pose, shape, and appearance in real human or KOL images, which can lead to misalignment during synthesis. To address this, GenKOL incorporates a pretrained facial landmark detection model, which identifies 68 key facial landmarks. By establishing a standardized initial pose, the system significantly improves the coherence and alignment of generated faces within the pipeline. The synthesized faces are then seamlessly reintegrated into their original context, ensuring both stability and visual quality. Figure~\ref{fig:kol_face_extractor} provides an example of a pre-aligned facial input used in the GenKOL workflow.

\subsection{System Architecture}

We propose a modular, plugin-based architecture that simplifies the integration of AI models as standalone services (Figure~\ref{fig:proposed_system}). Each service, such as image editing, text-to-image generation, or background replacement, is encapsulated as an independent module with a standardized interface, enabling seamless communication and sequential execution. This design allows services to be installed, replaced, or removed with minimal effort, supporting rapid experimentation with minimal resource overhead. By loading only the necessary components, the system optimizes resource use and improves scalability, while support for multiple algorithmic versions enables users to balance speed, accuracy, and visual quality depending on application needs. The architecture consists of four key components: Engine, Flow Generator, Services, and Algorithms.

\textbf{Engine. }The Engine acts as a proxy layer that standardizes communication among AI services through a uniform interface. All services, whether deployed locally or in the cloud (e.g., AWS, Google Cloud), must conform to this interface to be registered in the system. This plug-and-play design not only streamlines integration and simplifies maintenance but also enables dynamic reconfiguration of workflows without disrupting overall functionality. Such flexibility reduces the technical burden for users and accelerates the deployment of new capabilities.

\textbf{Flow Generator. }The Flow Generator composes executable pipelines from registered services, allowing users to construct task-specific workflows. It supports iterative refinement of AI-driven processing chains, enabling the combination of services such as garment transfer and makeup application into cohesive pipelines. By automating pipeline construction, the Flow Generator reduces configuration errors, minimizes manual intervention, and saves time for both developers and end-users.

\textbf{Services. }The Service layer manages the registration and execution context of algorithms, whether implemented as local machine learning models or remote inference APIs. This design gives users fine-grained control over resource allocation, enabling deployment decisions based on available memory, GPU capacity, or scalability requirements. As a result, the system can adapt to diverse computational environments, from lightweight personal devices to large-scale commercial infrastructure, making it cost-effective and efficient.

\textbf{Algorithms. }Each service (e.g., virtual dressing, makeup transfer, or scene editing) can employ multiple interchangeable algorithms. These algorithms are deployment-agnostic and can be executed across heterogeneous environments, including local devices, private servers, and cloud-based platforms. This level of flexibility ensures that the architecture can accommodate a wide range of use cases, from individual content creation to enterprise-scale marketing campaigns.

\begin{figure*}[!t]
    \centering
    \includegraphics[width=\linewidth]{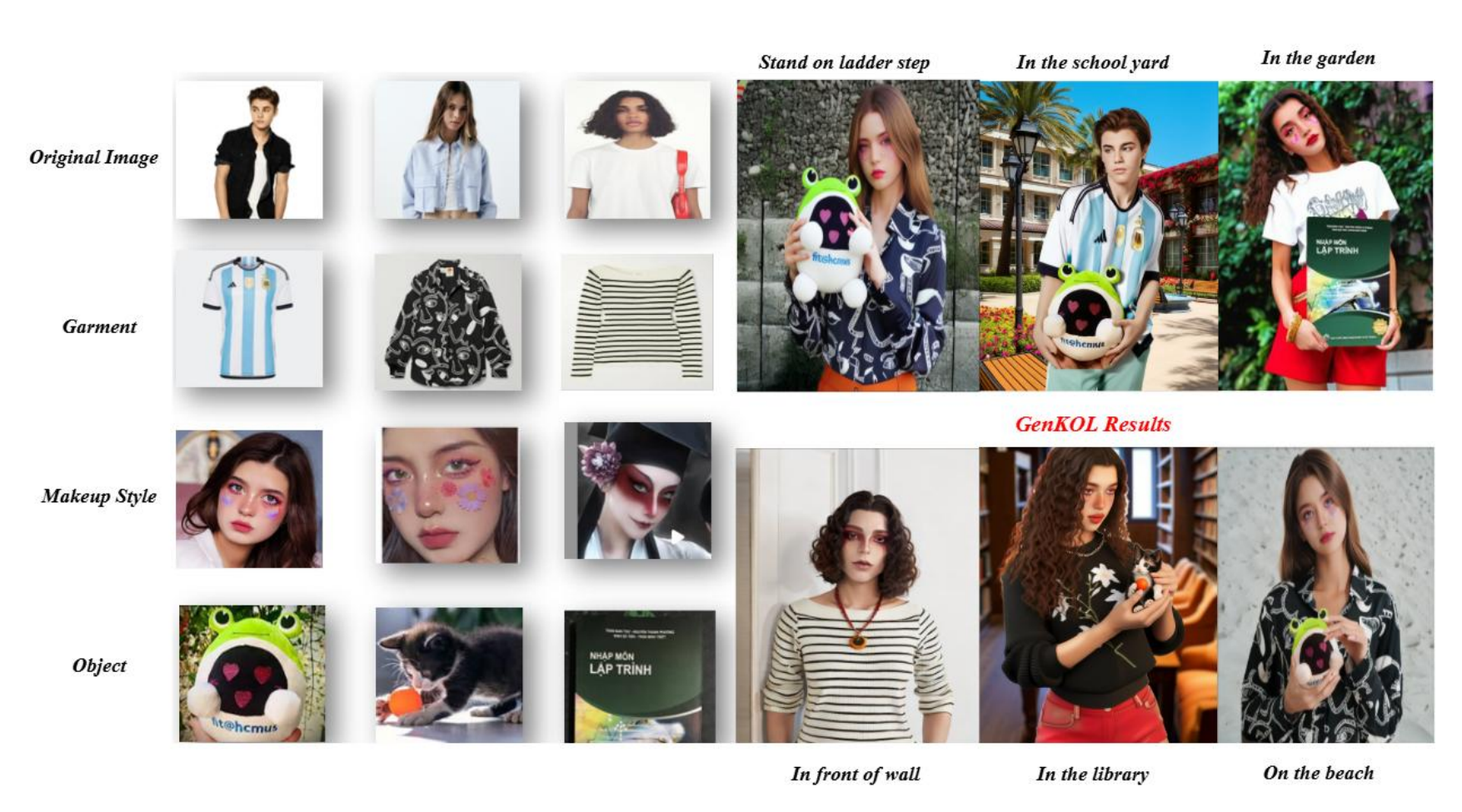}
    \caption{Examples of generated results of GenKOL. Given an original image (top row, left) and corresponding prompts for each attribute (garment, makeup style, and interaction object), our GenKOL system (rightmost column) synthesizes realistic virtual KOLs that seamlessly combine all elements, including the specified background.}
    \label{fig:our_result}
\end{figure*}

\subsection{Modular Extensibility}

GenKOL is designed with a plugin-based architecture that facilitates seamless integration of new models. To incorporate a new model, users first register it with the Engine, which functions as a centralized controller and proxy for all services. This registration step enables the model to be referenced, scheduled, and invoked consistently throughout the system.

Following registration, users implement the corresponding service logic within the Algorithms module. This includes defining standardized input and output interfaces as well as specifying the execution method tailored to the model’s task (e.g., garment synthesis, background replacement, or facial attribute editing). Once these steps are complete, the model is encapsulated as a plugin, allowing it to be dynamically inserted into any user-defined processing pipeline.

The Engine orchestrates execution by invoking the appropriate plugin services with the correct inputs. To guarantee correct execution order across interdependent services, a dependency matrix is maintained to represent relationships among plugins. This structured approach allows the system to automatically determine valid execution sequences when constructing pipelines, minimizing manual coordination and reducing error propagation.

An example of the plugin integration workflow, including model registration, method binding, and service mapping, is illustrated in Figure~\ref{fig:modular-extend}. This extensibility ensures that GenKOL can readily adapt to advances in generative AI by supporting the rapid incorporation of emerging models without disrupting existing workflows.

%% file: content/exp.tex
\section{Experiments}
\label{experiments}

\begin{figure*}[!t]
    \centering
    \includegraphics[width=\linewidth]{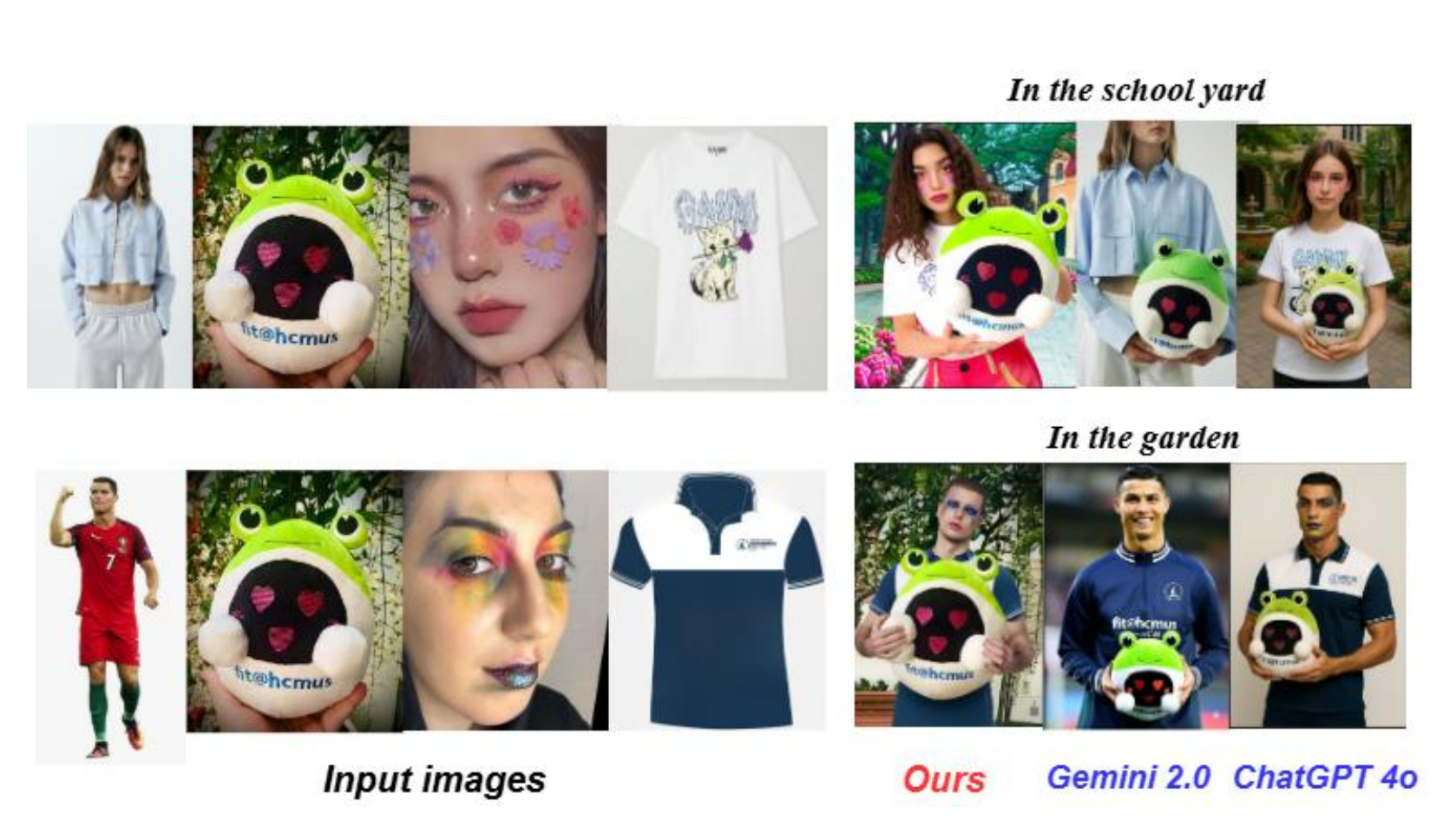}
    \caption{Qualitative comparison of image outputs from GenKOL (Ours), Gemini-2.0, and ChatGPT-4o across diverse scenes and visual prompts.}
    \label{fig:compared}
\end{figure*}

\subsection{Experiments Settings}
\textbf{Since no current application provides the complete range of functionalities that GenKOL does, direct comparisons on a one-to-one basis are not possible.} Most tools concentrate on a specific feature; for example, KlingAI\footnote{\url{https://www.klingai.com/global/}} focuses on facial editing and makeup effects, while Fitroom\footnote{\url{https://fitroom.app}} facilitates virtual clothing try-ons but does not include facial features. Maybelline\footnote{\url{https://www.maybelline.com/virtual-makeover-makeup-tools}} allows for cosmetic try-ons without integrating clothing or full image generation, and TRYO\footnote{\url{https://apps.apple.com/us/app/tryo-virtual-try-on-ar-app/id1640247631}} offers AR-based try-ons but lacks AI-driven generation capabilities. In contrast, GenKOL consolidates these functionalities, allowing for the intuitive and efficient creation of customizable virtual Key Opinion Leaders (KOLs). A comparison of the functions available in various key tools alongside GenKOL is shown in Table \ref{tab:survey}.


\begin{table}[!t]
\caption{Survey of Available Features in Selected Tools and Applications Compared to GenKOL.}

\label{tab:survey}
\resizebox{\linewidth}{!}{%
\begin{tabular}{l|c|c|c|c}
\toprule
\textbf{Tool} & \textbf{Virtual Try On} & \textbf{Makeup} & \textbf{Background Modify} & \textbf{Object Interaction} \\ \midrule
{KlingAI}    & \xmark          & \cmark           & \cmark & \xmark \\
{Fitroom}    & \cmark          & \xmark & \xmark          & \xmark \\
{Maybelline} & \xmark & \cmark          & \xmark          & \xmark \\
{TRYO}       & \cmark          & \cmark          & \xmark          & \xmark \\
\rowcolor{lightgray} \textbf{GenKOL}     & \cmark          & \cmark           & \cmark           & \cmark  \\
\bottomrule
\end{tabular}%
}
\end{table}

In the absence of a comprehensive benchmark, we evaluated GenKOL through user satisfaction studies to obtain impartial feedback and assess the effectiveness of GenKOL as an intelligent system for image generation. The experiments were executed on a Linux server with 40 GB NVIDIA GPUs to facilitate parallel image generation during the evaluation phase.

\subsection{Qualitative Evaluation}

\subsubsection{Showcase of Generated Images}

\begin{table}[!t]
    \centering
    \renewcommand{\arraystretch}{1.4}
    \caption{Qualitative comparison between GenKOL, Gemini-2.0, and ChatGPT-4o in terms of image generation time, perceptual quality, and object consistency.}
    \resizebox{\linewidth}{!}{%
    \begin{tabular}{l|c|c|c}
        \toprule
        \textbf{Method} & 
        \textbf{Average Generation Time (s)} & 
        \textbf{Image Quality} & 
        \textbf{Consistency} \\
                \midrule
        ~Gemini-2.0 & \textbf{30} & Neutral & Bad \\
        ChatGPT-4o & 600 & \textbf{Very Good} &  Neutral\\
        GenKOL & 300 & Good & \textbf{Very Good} \\
        \bottomrule
    \end{tabular}
    } \label{tab:model_comparison}
\end{table}


The findings from our experimental image generation demonstrate the capability of GenKOL in producing highly realistic advertising visuals. The generated product images showcased a variety of styles, genders, ages, and contexts, providing numerous options to satisfy the needs of any marketing initiative. Figure~\ref{fig:our_result} illustrate examples of these KOL images, which boast remarkable quality and realism. This breakthrough greatly minimizes the time required for product and brand design, leading to significant cost savings for companies. As a result, organizations can better allocate their resources to areas such as product development and customer engagement, improving the overall effectiveness of marketing campaigns.

\begin{figure*}[!t]
    \centering    \includegraphics[width=\linewidth]{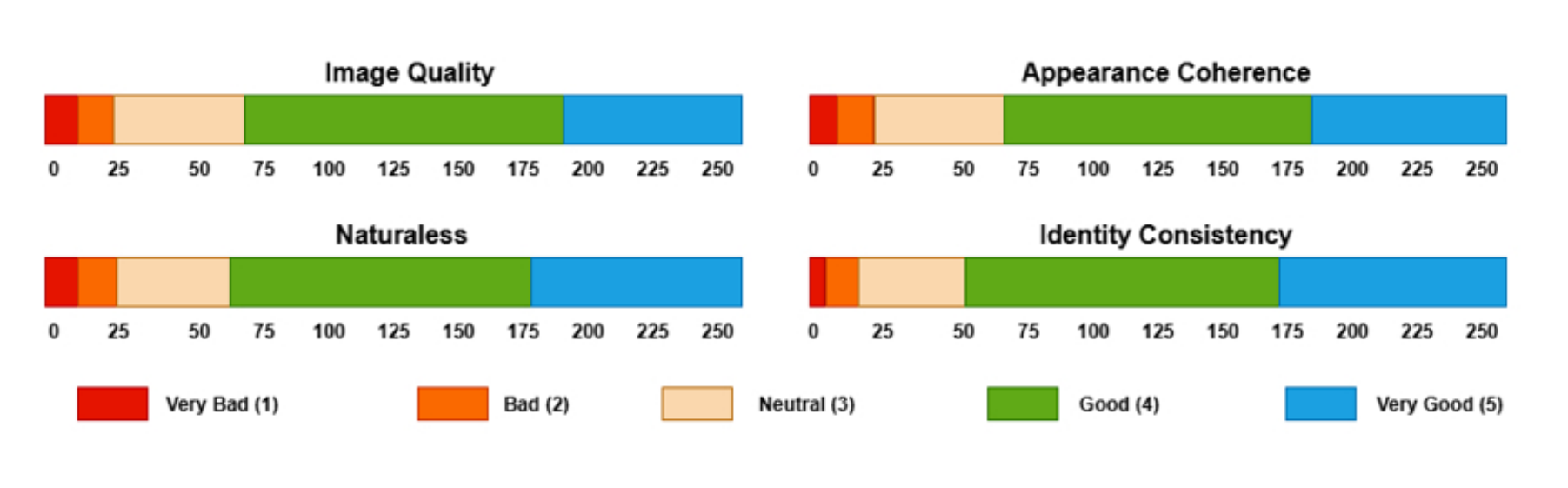}
    \vspace{-15pt}
    \caption{
    Rating distributions across four evaluation metrics in the user study of generated images. The horizontal bars indicate the aggregated scores for each metric.
    }
\label{fig:result_evaluation}
\end{figure*}

\begin{figure*}[!t]
    \centering
    \includegraphics[width=\linewidth]{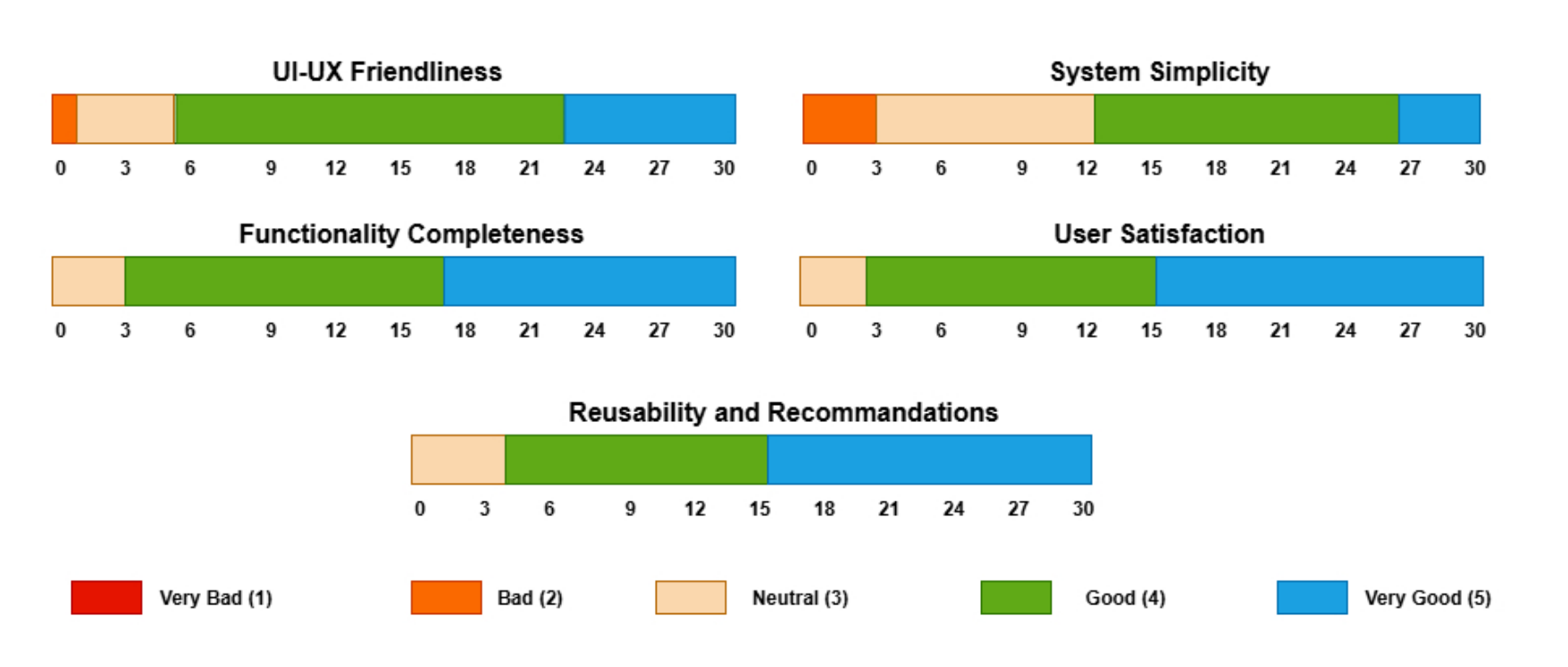}
    \caption{User study rating distributions for the proposed GenKOL system across five evaluation metrics: UI–UX Friendliness, System Simplicity, Functionality Completeness, User Satisfaction, and Reusability \& Recommendations.}
    \label{fig:exp_all_pie}
\end{figure*}

\subsubsection{Comparison with Gemini-2.0 and ChatGPT-4o}

We evaluated the GenKOL system alongside state-of-the-art image generation models, conducting a controlled comparison with Gemini-2.0-Flash-Preview-Image-Generation and ChatGPT-4o. The assessment focused on three key criteria: image generation time, visual quality, and contextual consistency (Figure~\ref{fig:compared}). Experimental results show that GenKOL produces high-quality images with strong consistency in clothing, makeup, and environmental interactions. Although GenKOL requires longer generation times compared to Gemini-2.0-Flash-Preview-Image-Generation, its outputs exhibit richer detail and superior contextual coherence. In contrast, GenKOL achieves generation speeds nearly twice as fast as ChatGPT-4o, while delivering only slightly lower visual quality. A detailed comparison of performance across models, including both quantitative and qualitative metrics, is presented in Table~\ref{tab:model_comparison}.

\subsection{User Study}

\subsection{Evaluation of Generated Images}

To evaluate the effectiveness of image generation, we conducted a comprehensive user study focusing on both the visual outcomes and the prompts used during generation. A total of 254 participants, aged 15–35, were recruited from diverse professional fields, including software development, economics, engineering, and higher education. The objective was to obtain insights into the practical usability of the system, user perceptions, and overall satisfaction.

We compiled a dataset of 200 high-quality virtual KOL images and organized the evaluation into four themed Google Forms, each targeting a different functional aspect of the system: Try-On, Makeup Application, Background Replacement, and Object Interaction. Each form provided step-by-step instructions, illustrative examples, and carefully curated image sets for assessment.

Participants rated the generated images on a 5-point Likert scale (1 = Very Bad, 5 = Very Good) according to factors such as realism, relevance, and visual quality. To complement the quantitative assessment, open comment boxes allowed participants to provide qualitative feedback at the end of each form.

The results, summarized in Figure~\ref{fig:result_evaluation}, indicate high levels of user satisfaction, with most ratings concentrated in the 4–5 range (Good to Very Good). Feedback highlighted GenKOL’s ability to deliver realistic, aesthetically appealing, and contextually relevant visuals. These findings confirm the system’s strong potential for generating marketing-ready content that meets or exceeds user expectations in terms of realism, fidelity, dimensionality, and aesthetic quality.

\subsection{Evaluation of GenKOL System}

In addition to assessing the quality of visual outputs, we conducted a user experience (UX) study to evaluate the friendliness, usability, and accessibility of the GenKOL platform. To ensure broad applicability, an open survey was administered to 30 participants from diverse professional fields, including economics, engineering, and non-IT sectors.

Three different generation pipelines were designed for participants to interact with, ranging from simple, single-service transformations to more complex workflows involving multiple services such as makeup application, garment transfer, and object interaction. To facilitate the process, five curated sample input images were provided for each service, while participants were also encouraged to use self-selected images from online sources. This dual approach ensured both consistency in evaluation and flexibility in exploring the system’s adaptability across varied contexts.

Participants were instructed to explore the specific features of GenKOL and evaluate their experiences on a 5-point Likert scale (1 = Very Bad, 5 = Very Good). The assessment covered five criteria: UI/UX Friendliness, System Simplicity, Functionality Completeness, User Satisfaction, and Reusability and Recommendations. This structured design allowed us to capture both the accessibility of the interface and the practicality of the multistep generation workflows.

As shown in Figure~\ref{fig:exp_all_pie}, results indicate consistently positive evaluations, with the majority of ratings falling within the 4–5 range across all five criteria. These findings confirm that GenKOL provides an intuitive, user-friendly interface and supports high-quality, adaptable virtual KOL generation, demonstrating strong usability and accessibility for end users.

%% file: content/conclusion.tex
\section{Conclusions and Future Work}
\label{conclusion}

In this paper, we present a modular, plugin-based architecture designed to simplify the integration of diverse AI services, particularly GenKOL. GenKOL creates realistic visuals of virtual KOLs and is recognized for its flexibility, scalability, and user-friendly design, making it ideal for AI-driven visual content in marketing and e-commerce. It enables dynamic workflows for tasks such as virtual dressing and makeup applications, significantly reducing time, resources, and production costs while ensuring high-quality output. 

However, GenKOL faces challenges that we plan to address in future updates. The quality of the output depends on the performance and compatibility of individual plugins, prompting us to develop an automated plugin validation system. We also aim to optimize pipeline execution by adaptively reordering tasks to enhance performance. These improvements will solidify GenKOL’s position as a reliable platform for next-generation AI-driven visual content generation.